\documentclass[prologue, table, sigconf]{acmart}

\setcopyright{rightsretained}

\usepackage{pgfplotstable}
\usepackage{filecontents}
\usepackage{multirow}
\usepackage{algorithm}
\usepackage{algpseudocode}
\usepackage{subcaption}
\usepackage{xcolor}

\begin{document}

\title{The Effect of Epigenetic Blocking on Dynamic Multi-Objective Optimisation Problems}

\acmConference[GECCO '22 Companion]{The Genetic and Evolutionary Computation Conference}{July 9--13}{Boston, MA, USA}

\author{Sizhe Yuen}
\orcid{1}
\affiliation{%
  \institution{University of Southampton}
  \department{Maritime Engineering}
  \city{Southampton}
  \country{United Kingdom}
}
\email{s.yuen@soton.ac.uk}

\author{Thomas H.G. Ezard}
\orcid{1}
\affiliation{%
  \institution{National Oceanography Centre}
  \department{Ocean and Earth Science}
  \city{Southampton}
  \country{United Kingdom}
}
\email{t.ezard@soton.ac.uk}

\author{Adam J. Sobey}
\orcid{1}
\affiliation{%
  \institution{The Alan Turing Institute}
  \department{Marine and Maritime Group, Data-centre Engineering}
  \city{London}
  \country{United Kingdom}
}
\email{ajs502@soton.ac.uk}


\begin{abstract}
  Hundreds of Evolutionary Computation approaches have been reported.
  From an evolutionary perspective they focus on two fundamental
  mechanisms: cultural inheritance in Swarm Intelligence and genetic
  inheritance in Evolutionary Algorithms. Contemporary evolutionary biology looks beyond
  genetic inheritance, proposing a so-called ``Extended
  Evolutionary Synthesis''. Many concepts from the Extended Evolutionary Synthesis have been 
  left out of Evolutionary Computation as interest has moved towards
  specific implementations of the same general mechanisms. One such
  concept is epigenetic inheritance, which is increasingly considered central
  to evolutionary thinking. Epigenetic mechanisms allow quick non- or
  partially-genetic adaptations to environmental changes. Dynamic
  multi-objective optimisation problems represent similar circumstances
  to the natural world where fitness can be determined by multiple objectives
  (traits), and the environment is constantly changing. 

  This paper asks if the advantages that epigenetic inheritance provide
  in the natural world are replicated in dynamic multi-objective optimisation problems. 
  Specifically, an epigenetic blocking mechanism is applied to a
  state-of-the-art multi-objective genetic algorithm, MOEA/D-DE, and its
  performance is compared on three sets of dynamic test functions, FDA, JY,
  and UDF. The mechanism shows improved performance on 12 of the 16 test
  problems, providing initial evidence that more algorithms should explore
  the wealth of epigenetic mechanisms seen in the natural world.
\end{abstract}

\ccsdesc[500]{Theory of computation~Evolutionary algorithms}
\ccsdesc[500]{Computing methodologies~Genetic algorithms}

\keywords{Genetic algorithms, multi-objective optimisation, dynamic optimisation, epigenetics}

\maketitle

\section{Borrowing from the Extended Evolutionary Synthesis}
The Modern Synthesis \cite{modern-synthesis}, a combination of Darwin and Wallace's
ideas of natural selection \cite{darwin-wallace, darwin:1859},
and Mendel's principles of inheritance \cite{mendel}, has been an important
inspiration for the concepts used in Evolutionary Algorithms.
Despite the number of approaches available,
the core inspiration is often just genetic inheritance.
However, modern evolutionary theory has since continued to explore
the mechanisms of evolution, extending the Modern Synthesis
to include concepts of non-genetic inheritance such as epigenetics, parental effects,
multilevel selection, and cultural inheritance in a portfolio proposed as the Extended Evolutionary Synthesis
\cite{pigliu-extended-synthesis, extended-synthesis-book}.



Epigenetic mechanisms in evolutionary theory alter DNA expression,
leading to a change in phenotype without a change in the underlying genotype
\cite{epigenetic-mice}. This ``phenotypic plasticity'' leads to a faster rate of change to quickly
adapt to changes in the environment, and the ability to revert changes if
the environmental conditions do not activate the epigenetic mechanism.
This is to allow adaptation to a natural world that is changing, optimising organismal fitness
without altering its underlying genotype. Dynamic problems reflect a similar set of challenges by
changing the optimal Pareto set or Pareto front over time \cite{fda}.
This paper therefore explores if, and, if so, how epigenetics might be used within
Evolutionary Algorithms to improve performance. Due to the faster-than-generational
adaptation capabilities of epigenetics, these are first
explored in a dynamic multi-objective problem context.

There is limited literature exploring epigenetic mechanisms in
genetic algorithms; those that exist are focused on static single-objective problems. 
Results are nevertheless consistent with expectations based on evolutionary theory.
The Knapsack problem was used to determine the performance of
the cytosine methylation epigenetic process when added to a traditional
genetic algorithm \cite{chrominski-ega}, resulting in a 25-30\% reduction
in the number of generations needed to reach an optimal solution.
A constant probability of methylation is used,
so the rate at which genes are blocked do not react to
dynamic changes based on external factors or progression
of the search. In epigenetic models, constant rates
of variation are
non-specific to the environment and act similar to mutation,
while varying rates are
directed by environmental cues \cite{jablonka-epimodels}.


\section{Epigenetic blocking mechanism}
\begin{figure}[H]
  \centering
  \includegraphics[width=0.47\textwidth, keepaspectratio]{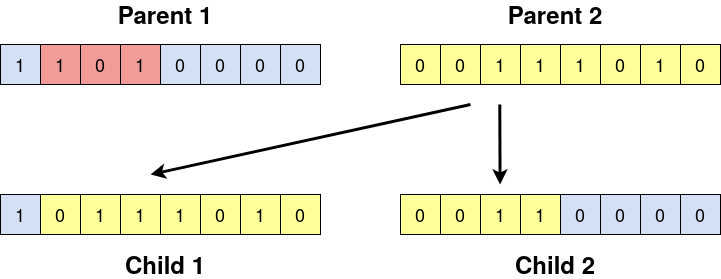}
  \caption{The blocking mechanism where some variables
    are blocked from carrying over to the next generation.}
  \label{fig:methylation-ga}
\end{figure}

There are three forms of epigenetic transfer possible: mitotic, germline,
and experience-dependent \cite{arabadopsis-colot, champagne-rodents}.
Germline and experience-dependent transfer passes down epigenetic marks
that direct the epigenetic process for future generations, while
mitotic transfer only propagates changes in the same generation.
For an initial exploration of epigenetic mechanisms, a simple form
of genetic blocking is chosen for inspiration based on mitotic transfer.
A probabilistic blocking mechanism is used to
block some variables in each individual from being changed during crossover.
The parameters of the epigenetic process are not inherited across
multiple generations (Figure \ref{fig:methylation-ga}). A simple mechanism
allows the properties to be analysed and to more clearly understand the effect
of an epigenetic process on the performance of an algorithm. The simplicity
also allows the mechanism to be adapted to any Evolutionary Algorithm
by altering the crossover method without adding complicated features.

The mechanism has a probability to trigger
during the reproduction stage for every parent without bias towards fitness as blocking both 
the fitter and less-fit parents have merits. Blocking fitter parents reduces
stagnation of the population, i.e. maintains diversity, should the variables or objectives of the dynamic
problem changes, while blocking less fit parents increases the convergence of
the population through more rapid selection at the variable level.

%
%

By varying the probability and the number of genes
that are blocked, and controlling the duration of each dynamic cycle
in a problem, the impact of epigenetic blocking can be analysed
and compared to a baseline algorithm, and to a constant probability
of triggering the mechanism.

\section{Experimental setup}
The MOEA/D-DE algorithm \cite{moead-de} with the
re-initialisation strategy outlined in \cite{udf}
was chosen as the base algorithm for benchmarking due to its strong performance
on dynamic optimisation problems. The fast converging nature of MOEA/D lets it
adapt quickly in changing environments \cite{jy}.
A population size of 500 is used.

Three variants of the epigenetic blocking mechanism are investigated by changing two parameters:
the probability for the mechanism to trigger, and the number
of variables blocked in the process (block size). A summary of the three variants
is as follows:
\begin{itemize}
\item \textbf{E} - with a constant probability of 0.1 and
  a constant block size of 6.
\item \textbf{EIB} - with a constant probability of 0.1
  and a varying block size from 1 up to the number of variables
  in the problem.
\item \textbf{EIP} - with a varying probability from 0
  up to a maximum of 0.8 with a constant block size.
\end{itemize}

The gradual increase in either
probability or block size is intended to increase
the convergence of the population,
as blocking more prevents diverse changes. The maximum
probability is limited to 0.8 to prevent stagnation
where the blocking occurs too often.
Increases to probability are rounded to 0.01,
increasing every 2 generations to the maximum probability.
The block size increases are rounded to the nearest whole number and
depends on the number of variables. For example, a problem with 30 variables
and a population of 500 will increase the block size every 4 generations.

\begin{table}
\caption{Hyperparameters table}
\label{tbl:hyperparameters}
\begin{center}
  \begin{tabular}{c | c}
    \textbf{Algorithm} & \textbf{Hyperparameters} \\ \hline
    MOEA/D-DE & $P(m)=1/D$, $P(x)=0.9, CR = 1, F = 0.5$\\
    E & $P(b) = 0.1, s = 6$ \\
    EIB & $P(b) = 0.1, s = \frac{evals}{max\_evals} * num\_vars$ \\
    EIP & $P(b) = \frac{evals}{max\_evals} * 0.8, s = 6$
  \end{tabular}
\end{center}
\end{table}

Details of these hyperparameters are shown in in Table \ref{tbl:hyperparameters}.
$P(b)$ is the
probability for the mechanism to trigger and $s$ is the block size.
The Inverted Generational Distance (IGD) \cite{igd} metric is used
to show the performance of the epigenetic variants to the MOEA/D-DE
algorithm.


The FDA \cite{fda}, JY \cite{jy}, and UDF \cite{udf} benchmark functions are
chosen to test the performance of the epigenetic blocking mechanism.
The FDA and UDF functions are considered to be simpler to solve as they are
based on the multi-objective ZDT \cite{zdt} problems while the JY
problems are more complex \cite{sdp}, including elements such as linkage between
variables and multiple knee points. For the benchmark
problem properties, $\tau$ is set at 5 and $nT$ at 10. This gives 5 generations
before the problem changes and 10 distinct steps. In total this gives
100 generations (50,000 iterations) to complete a full cycle back to the
original variables and objectives of the problem. 2 full
cycles are benchmarked with 20 independent runs for analysis.

\section{Performance on dynamic multi-objective optimisation problems}
To determine how the changes in the block rate and the probability of
change affect the performance, the three variants are compared by
taking the average IGD every generation to match the rate
at which the probability varies.
The difference between each variant and the
baseline are then compared to demonstrate the improvement
in performance.

\subsection{Performance of epigenetic approaches against each other}

\definecolor{good}{HTML}{87CEEB}

\newcommand{\good}{\cellcolor{good!50}}
\newcommand{\bad}{\cellcolor{red!60}}

\begin{table}
  \caption{The total \% difference for a two full dynamic cycles. The p-values
    from a Wilcoxon signed-rank test is shown in brackets, bold indicates the best performing
    variant. The lighter blue boxes indicate an improvement in performance;
    darker red boxes indicate a decrease.}
  \label{tbl:diff-sum}

  \resizebox{0.48\textwidth}{!}{
  \begin{tabular}{c | c | c | c}
    \textbf{Problem} & \textbf{E} & \textbf{EIB} & \textbf{EIP}             \\
    \hline
    FDA1 & \bad -102          (0.000) & \bad -14            (0.027) & \bad -188          (0.015) \\ 
    FDA2 & \good 399          (0.000) & \good \textbf{400}  (0.000) & \good 316          (0.000) \\ 
    FDA3 & \good 163          (0.000) & \good 189           (0.016) & \good \textbf{810} (0.000) \\
    \hline
    JY1 &  \good 657          (0.000) & \good 504           (0.000) & \good \textbf{917} (0.000) \\ 
    JY2 &  \good 193          (0.012) & 41                  (0.783) & \good \textbf{470} (0.000) \\ 
    JY3 &  \good 669          (0.000) & \good \textbf{1164} (0.000) & \good 422          (0.000) \\ 
    JY5 &  \bad -231          (0.000) & -28                 (0.518) & \good \textbf{226} (0.002) \\ 
    JY6 & -19                 (0.209) & -15                 (0.164) & \bad -30           (0.020) \\ 
    JY7 & 96                  (0.871) & \good 318           (0.029) & \good \textbf{391} (0.001) \\ 
    JY8 & 16                  (0.108) & \good 29            (0.028) & \good \textbf{37}  (0.001) \\
    \hline
    UDF1 & \bad -181          (0.031) & \good \textbf{170}  (0.013) & -35                (0.687) \\ 
    UDF2 & -52                (0.323) & 15                  (0.728) & \bad -40           (0.024) \\ 
    UDF3 & \good \textbf{61}  (0.000) & \good -17           (0.000) & 35                 (0.586) \\ 
    UDF4 & \good 124          (0.003) & \good \textbf{413}  (0.000) & \bad -71           (0.027) \\
    UDF5 & -2                 (0.217) & 122                 (0.262) & -120               (0.287) \\
    UDF6 & \good \textbf{206} (0.000) & \good 180           (0.000) & \good 128          (0.000) \\
  \end{tabular}}
\end{table}

The total percentage difference between the baseline
and each variant are summarised in Table \ref{tbl:diff-sum}.
Positive values represent the epigenetic mechanism performing
better than the baseline algorithm and
negative values a decrease. Values highlighted have a p-value
lower than 0.05 from a Wilcoxon signed-rank test
and are statistically significant at the 95\% level.

The performance summary of the three variants is as follows:
\begin{itemize}
\item \textbf{E} - Positive performance on 8 out of 16 problems, best on 2 problems.
\item \textbf{EIB} - Positive performance on 10 out of 16 problems, best on 4 problems.
\item \textbf{EIP} - Positive performance on 9 out of 16 problems, best on 6 problems.
\end{itemize}

The presence of epigenetic blocking generally improves the performance
compared to the baseline. It is only the FDA1 case where all of the
epigenetic enhanced algorithms struggle.
Both EIB and EIP generally outperform the E variant, where
the blocking rate and probability of blocking remain static. This is
expected as using a constant probability and block size does not allow the
mechanism to react dynamically to changes in the problem.

EIP performs best on the most problems, 6, with 5 of these 6 problems in
the JY problem set. The performance suggests that
EIP works best on more complex problems with its aggressive
use of the epigenetic mechanic compared to EIB and E. However, EIP
doesn't perform well on the UDF problem set, considered to be the simplest
problems that are usually dominated by convergence, with 3 non-detectable
results and 2 negative results where it performs worse.
The aggressive behaviour of this variant leads to
larger positive and negative results making it less robust.

In comparison, the EIB variant shows more consistency across the
problem sets with only 1 significant negative result and is able to achieve
a best result on each problem set.
Although EIP can find stronger positive results in some problems,
its inconsistency and poor performance on the UDF problem set shows that
it is more suitable for complex problems, whereas EIB finds positive
improvement across a larger range of problems. Both of these approaches
outperform the E mechanism, indicating that increasing blocking over
the lifecycle was beneficial.


\subsection{Performance against the baseline}
\begin{figure*}
  \centering
  \begin{subfigure}{0.3\textwidth}
    \includegraphics[width=\linewidth]{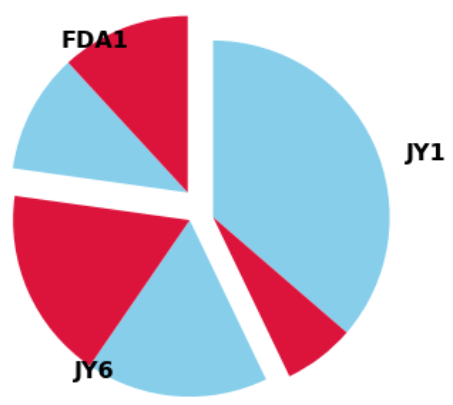}
    \subcaption{Category I problems}
  \end{subfigure}
  \hfill
  \begin{subfigure}{0.3\textwidth}
    \includegraphics[width=\linewidth]{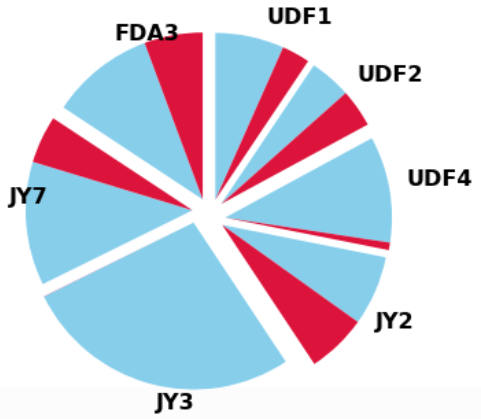}
    \subcaption{Category II problems}
  \end{subfigure}
  \hfill
  \begin{subfigure}{0.3\textwidth}
    \includegraphics[width=\linewidth]{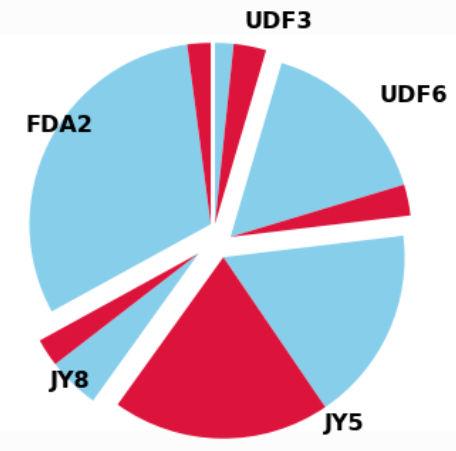}
    \subcaption{Category III problems}
  \end{subfigure}
  \caption{Comparison of EIB against the baseline MOEA/D-DE algorithm}
  \label{fig:eib-overview}
\end{figure*}

The total difference in performance between the baseline
and EIB over the 2 cycles is shown in Figure
\ref{fig:eib-overview}. The difference
on each two generation interval is summed with dark red
denoting when the baseline performs better and light blue denoting
when EIB performs better. In problems
such as JY5 and JY6, both algorithms have parts
of the search where they perform well, leading to
no detectable difference as can be seen in the p-value
in Table \ref{tbl:diff-sum}.
In problems with statistically significant negative performance
such as FDA1, it is not overwhelmingly negative,
with a positive improvement 40\% of the time. The
results show strong performance overall against the baseline,
indicating the strength of the epigenetic mechanism
in solving dynamic problems.

\section{Discussion}
The addition of an epigenetic blocking mechanism is able to significantly
improve the performance of the MOEA/D-DE algorithm on a range of dynamic problems.

Both the EIB and EIP variants performed better on more problems
compared to E, suggesting that varying the rates at which the mechanism
triggers and the variables that are blocked is more effective than keeping
the parameters static. From evolutionary theory, this behaviour is
expected in a dynamic environment with epigenetic variations that
are guided rather than random \cite{jablonka-epimodels}.
The varying rates give more control over the
convergence and diversity of the populations.
The increased consistency from EIB compared to EIP shows
the effectiveness of blocking more variables, even at a slower
rate. Convergence is increased with more variables blocked, hence it
performs well in general and on convergence based problems like UDF.
EIP blocks fewer variables but more often, finding success when 
the correct variables are blocked, but worse performance on the
wrong variables. More diversity can be retained with EIP as only
a small number of variables are blocked compared to blocking
a majority of variables in EIB. The increased diversity would explain
EIP's better performance on the complex JY problems, and it would
be expected for EIP to perform well on complex real world problems
where diversity plays a larger role than convergence.

The epigenetic mechanism shows more sensitivity to Pareto set changes
because the variables are directly blocked. Better performance is observed
on Category II and III problems where there are Pareto front changes, and
worse performance on the Category I problem where there are only Pareto set
changes. A change in the Pareto front is analogous to a change in objective  (fitness optimum)
in the natural world, which epigenetic processes can more quickly adapt to.
A change in the Pareto set is more complicated and can be related to the concept
of plasticity \cite{price-plasticity}, where the environment influences
the developmental stage of phenotypes.

An important advantage of this epigenetic mechanism is the ability to include
the mechanism into any crossover method of a genetic algorithm. The mechanism
sits on top of the crossover method, altering it slightly by blocking
the variables. In the natural world, epigenetics also sits on top of
existing genetic mechanisms to allow for quick adaptation.

\section{Conclusion}

This paper demonstrates the impact of including a simple epigenetic
blocking mechanism into an existing evolutionary algorithm, and
the advantages epigenetics can provide to algorithms solving
dynamic problems. The epigenetic mechanism improves upon the
baseline MOEA/D-DE algorithm on 12 of the 16 dynamic multi-objective
test problems, with a conclusively negative result on only 1 problem.
Increasing the number of variables blocked helps increase
convergence, giving consistent improvement on all categories
of test problems. Increasing the rate of which the mechanism
is activated is more effective on the complex JY problems, suggesting
a retention of diversity in the population.

There are further epigenetic mechanisms and features to look into.
In nature, epigenetics enable fast adaptations
triggered by environmental cues, and the epigenetic method
itself is inherited by future generations. The next step in the proposed blocking mechanism
is to include environmental cues to direct how the mechanism should be triggered,
and a form of inheritance to direct which variables should be blocked.



\bibliography{references}

\end{document}